\documentclass[letterpaper]{article} 
\usepackage{aaai24}  
\usepackage{times}  
\usepackage{helvet}  
\usepackage{courier}  
\usepackage[hyphens]{url}  
\usepackage{graphicx} 
\usepackage{natbib}  
\usepackage{caption} 
\usepackage{algorithm}
\usepackage{algorithmic}
\usepackage{booktabs}
\usepackage{multirow}
\usepackage{amsmath}
\usepackage{amssymb}
\usepackage{appendix}

\usepackage{newfloat}
\usepackage{listings}
\DeclareCaptionStyle{ruled}{labelfont=normalfont,labelsep=colon,strut=off} 
\floatstyle{ruled}
\newfloat{listing}{tb}{lst}{}
\floatname{listing}{Listing}
\title{All Beings Are Equal in Open Set Recognition}
\author{
    Chaohua Li\textsuperscript{\rm 1}\textsuperscript{,\rm 2}\equalcontrib,
    Enhao Zhang\textsuperscript{\rm 1}\textsuperscript{,\rm 2}\equalcontrib,
    Chuanxing Geng\textsuperscript{\rm 1}\textsuperscript{,\rm 2},
    Songcan Chen\textsuperscript{\rm 1}\textsuperscript{,\rm 2}\thanks{Corresponding author.}
}
\affiliations{
    \textsuperscript{\rm 1}College of Computer Science and Technology, Nanjing University of Aeronautics and Astronautics\\
    \textsuperscript{\rm 2}MIIT Key Laboratory of Pattern Analysis and Machine Intelligence

    \{chaohuali, zhangeh, gengchuanxing, s.chen\}@nuaa.edu.cn
}

\begin{document}

\maketitle

\begin{abstract}
In open-set recognition (OSR), a promising strategy is exploiting pseudo-unknown data outside given $K$ known classes as an additional $K$+$1$-th class to explicitly model potential open space. However, treating unknown classes without distinction is \textit{unequal} for them relative to known classes due to the category-agnostic and scale-agnostic of the unknowns. This inevitably not only disrupts the inherent distributions of unknown classes but also incurs both class-wise and instance-wise imbalances between known and unknown classes. Ideally, the OSR problem should model the whole class space as $K$+$\infty$, but enumerating all unknowns is impractical. Since the core of OSR is to effectively model the boundaries of known classes, this means just focusing on the unknowns nearing the boundaries of targeted known classes seems sufficient. Thus, as a compromise, we convert the open classes from infinite to $K$, with a novel concept \textit{Target-Aware Universum} (TAU) and propose a simple yet effective framework \textbf{D}ual \textbf{C}ontrastive Learning with \textbf{T}arget-\textbf{A}ware \textbf{U}niversum (DCTAU). In details, guided by the targeted known classes, TAU automatically expands the unknown classes from the previous $1$ to $K$, effectively alleviating the distribution disruption and the imbalance issues mentioned above. Then, a novel \textit{Dual Contrastive} (DC) loss is designed, where all instances irrespective of known or TAU are considered as positives to contrast with their respective negatives. Experimental results indicate DCTAU sets a new state-of-the-art.
\end{abstract}

\begin{figure}[t]
\centering
\includegraphics[width= 0.9\columnwidth]{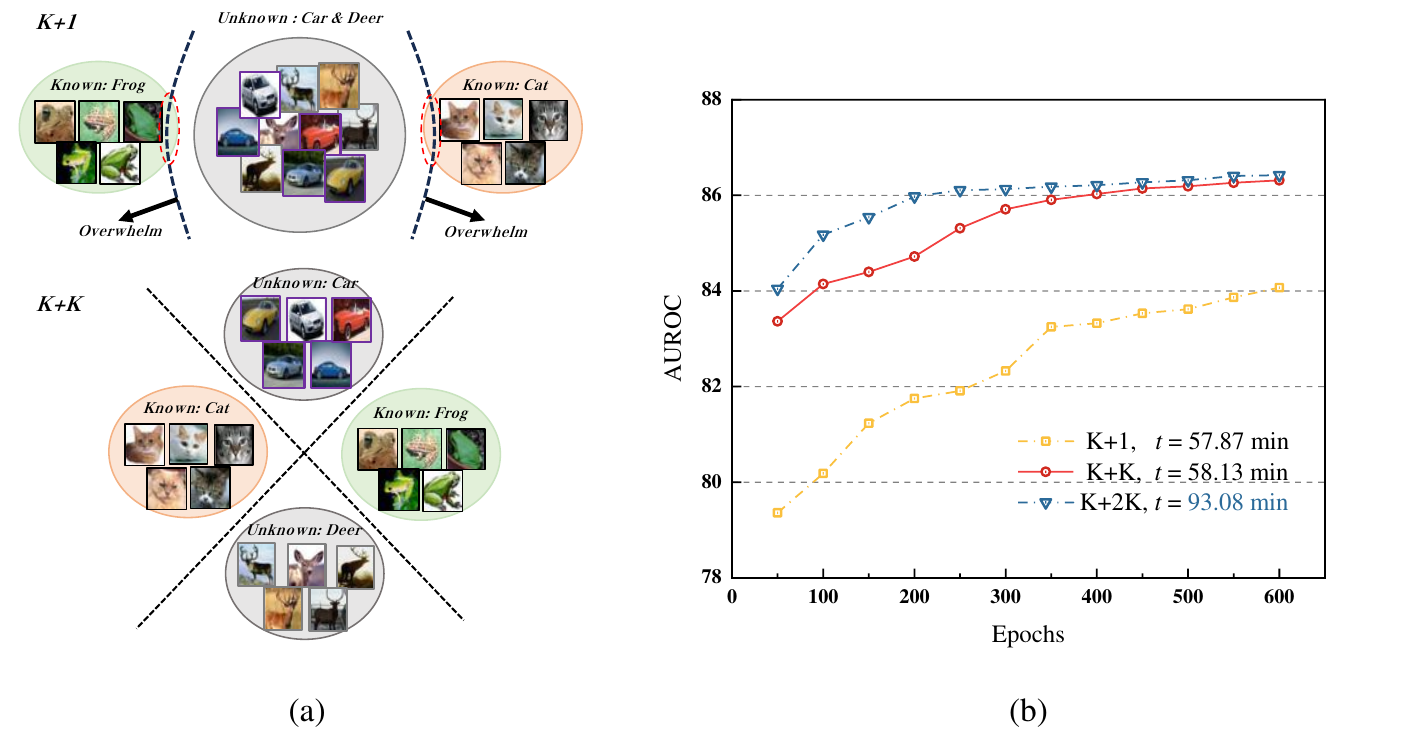} 
\caption{(a) The $K$+$1$ stereotype (\textit{top}) inevitably disrupts the inherent distributions of
unknowns and incurs a \textit{bigger} unknown class overwhelming other known classes. The $K$+$K$ strategy introduced (\textit{below}) can alleviate the issues existing in $K$+$1$; (b) An illustrated experiment on partial data of CIFAR10 indicates $K$+$K$ can be as a compromise. $K$+$K$ (\textit{red}) outperforms $K$+$1$ (\textit{yellow}) by a wide margin and shows comparable performance to $K$+$2K$ (\textit{blue}), while requiring less time cost.}
\label{fig1}
\end{figure}

\begin{figure*}[t]
\centering
\includegraphics[width=1.9\columnwidth]{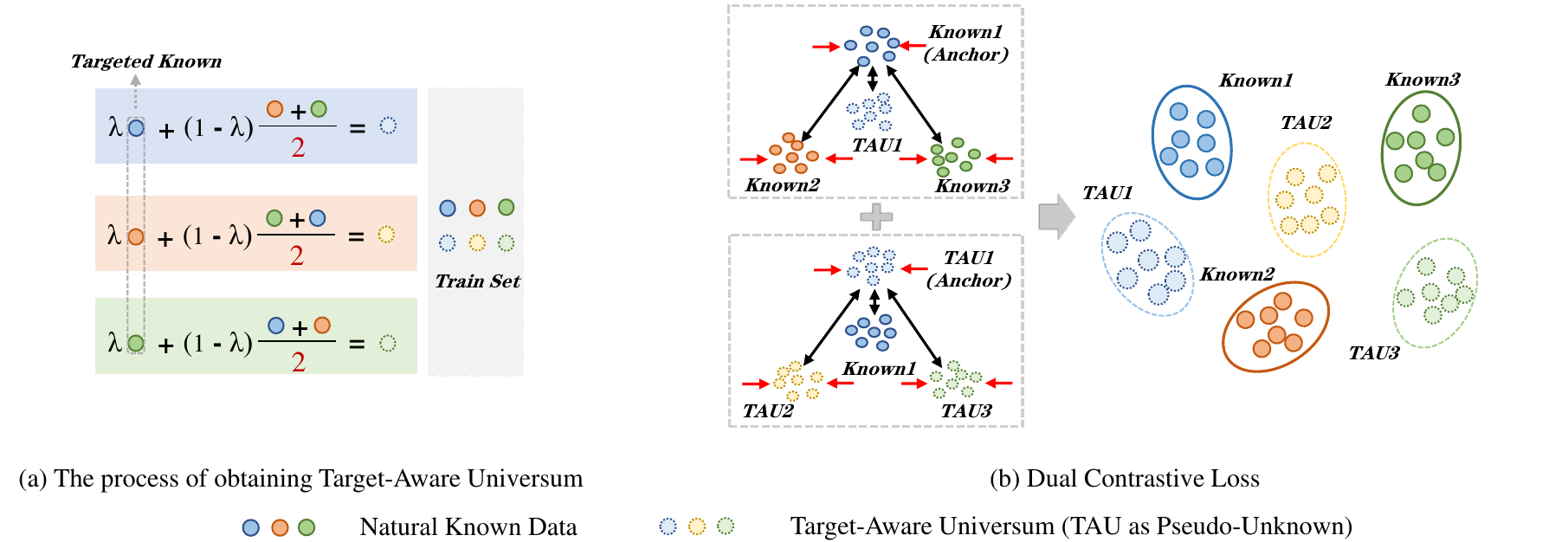} 
\caption{Two components of the proposed framework \textbf{D}ual \textbf{C}ontrastive Learning with \textbf{T}arget-\textbf{A}ware \textbf{U}niversum (DCTAU). (a) An illustration about how \textit{Target-Aware Universum}(TAU) is generated; (b) The \textit{Dual Contrastive} Loss is defined between the $i$-th targeted known class and other known classes \& the $i$-th TAU class (\textit{top}) and between the $i$-th TAU class and other TAU classes \& the $i$-th targeted known class (\textit{below}).}
\label{fig2}
\end{figure*}

\section{Introduction}
In real-world classification tasks, the deployed model may encounter data from unknown classes as the incomplete knowledge obtained during training\cite{geng2020recent}. Thus, a more realistic recognition scenario has been introduced, namely open-set recognition (OSR), aiming to classify known data and reject unknown data simultaneously \cite{scheirer2012toward}. 

With the rapid emergence of algorithms crafted for OSR, the strategy of utilizing pseudo-unknown data participated in training has garnered increasing attention, which involves obtaining the pseudo-unknown data through GANs or augmentation methods \cite{ge2017generative,neal2018open,kong2021opengan,du2022vos,zhou2021learning,dhamija2018reducing,gunther2017toward,perera2019deep,chen2021adversarial,xu2023contrastive}. This strategy treats the pseudo-unknown classes as an additional $K$+$1$-th class, excluding $K$ known classes, and improves performance to varying degrees. However, consider a realistic scenario, even at the first sight of animals and vehicles, we would never categorize them as one class without distinction at all. This reveals that it is counter-intuitive to treat the category-agnostic and scale-agnostic unknowns (\textit{i.e.,} the scale involves the number of instances in each unknown class and the number of classes of unknowns) as the $K$+1-th class. Such an extreme $K$+$1$ strategy will cause two issues. First, the inherent class distributions of unknowns will be disrupted since the initially specific categories of unknowns are fused into a single mixed class. Second, imbalance occurs on both class- and instance-wise: there are $K$ classes for knowns but only one class for unknowns; on the latter, copious unknown instances forced into a single mixed class will overwhelm the known classes, as illustrated in Fig. 1(a) (\textit{top}). The ideal (yet extreme) strategy should adopt the $K$+$\infty$, where modeling the whole class space involves $K$ classes of known classes and an infinite number of unknown classes. However, in reality, it is impractical to enumerate all unknown classes. To further clarify such two extremes, we design an illustrated experiment on partial data of CIFAR10. As shown in Fig.1 (b), with $K$+$1$ as a baseline (\textit{yellow line}), performance improves significantly as the number of pseudo-unknown classes grows to $K$ (\textit{red line}). Afterwards, with the further increase of it, performance plateaus and comes with substantial time consumption. For instance, when reached to $2K$, time consumption surprisingly rises by 60\% (\textit{blue line}). Therefore, we believe that using $K$ pseudo-unknown classes to approximate the $\infty$ open classes is a roughly reasonable option.

Given this, the challenge converts to how to generate $K$ pseudo-unknown classes. Since the core of OSR is recognized to effectively model the boundaries of known classes \cite{geng2020recent,vaze2021open}, here comes an intuition: as seen in Fig. 1(a) (\textit{below}), we should focus on generating $K$ pseudo-unknown classes for open classes nearing the boundaries of $K$ targeted known classes, which naturally converts to a $K$+$K$ strategy.

Upon the above, we first introduce a novel concept, \textit{Target-Aware Universum}, to serve as these $K$ pseudo-unknown classes. Then we design \textit{Dual Contrastive} loss to learn more discriminative representations around the boundaries between $K$ known classes and $K$ pseudo-unknown classes. We name this framework as Dual Contrastive Learning with Target-Aware Universum (DCTAU). Two main components of the framework are illustrated in Fig. 2. \textbf{(1) Target-Aware Universum} (TAU). The concept of TAU originates from the \textit{Universum Learning}, which introduces an external dataset that does not belong to any classes in the task \cite{weston2006inference,chapelle2007analysis}. To align perfectly with OSR task, TAU is the result of modifying the universum through \textit{Targeted Mixup}, which interpolates a targeted known and the average of the remaining knowns, as seen in Fig. 2(a). By this, each TAU nears to its targeted known since the targeted known contributes more information whereas the remaining known classes are averaged out. Meanwhile, TAU automatically expands the pseudo-unknowns from the previous 1 to $K$ classes which effectively alleviates the distribution disruption and the imbalance issues mentioned above. \textbf{(2) Dual Contrastive loss} (DC). To endow unknowns with the equal opportunity to be used as the positives, we design a novel DC loss, where all instances irrespective of known or TAU, aiming to simultaneously optimize greater inter-class margins and intra-class compactness for $K$ TAUs in a similar manner as $K$ known classes, as in Fig. 2(b). In DC loss, each anchor known class as the positive contrasts both other known classes and its TAU class firstly, meanwhile each TAU class contrasts both other TAU classes and its corresponding targeted known class. Further, theoretical analyses provide that DC loss can adaptively adjust the weights between knowns and TAUs in gradient update process to achieve fairly contrast and inherit the inner ability of hard negative mining. Ultimately, these two components enable DCTAU framework to achieve all beings (\textit{i.e.,} known and unknown classes) are equal in OSR. The contributions of this paper are summarized as follows:
\begin{itemize}
\item We emphasize the comprehension of unknowns and break the $K$+$1$ stereotype to a $K$+$K$ strategy, achieving all beings are equal in OSR.
\item We propose a novel framework Dual Contrastive Learning with Target-Aware Universum (DCTAU) involving Target-Aware Universum and  Dual Contrastive loss.
\item  We theoretically analyzed the effectiveness of DCTAU and extensive experiments show it surpasses the state-of-the-art performance.
\end{itemize}

\section{Related Work}
\subsection{Open Set Recognition}

\cite{scheirer2012toward} pioneered the formalization of open set recognition. \cite{bendale2016towards}  first integrates deep learning into the OSR task and proposed OpenMax. Subsequently, deep learning has gained significant attention in the OSR task. One of those focuses on transforming closed training into open. Apart from directly using natural images, it is broadly classified into two categories: 

\noindent \textbf{GANs Based Methods.} Seminal work in \cite{ge2017generative} extended Openmax, using a conditional GAN to synthesize unknown data to train the DNNs. Following this,  \cite{neal2018open,perera2020generative,kong2021opengan} enhanced the performance by different strategies based on GANs. Recently, \cite{chen2021adversarial} is an extension of \cite{chen2020learning} and adds confusing training samples from a generator. However, the training process of GANs is complex and unstable as it needs to train an additional network. 

\noindent \textbf{Augmentation Based Methods.} \cite{zhou2021learning} introduced Data Placeholder, which anticipates novel class patterns by Manifold Mixup \cite{verma2019manifold}. \cite{cho2022towards} used background-class as \textit{Known Unknown Classes} during training. \cite{xu2023contrastive} synthesized vague virtual instances and augmented negatives to enhance representation learning. \cite{zhu2023openmix} proposed OpenMix, which learned to reject pseudo data mixed through outlier samples. However, these methods not only consider the scale and quality of augmented instances but also treat the pseudo-unknown as the $K$+$1$-th class. 

\subsection{Contrastive learning}
Contrastive learning has become a dominant component in Self-Supervised Learning (SSL) \cite{duan2018deep,hendrycks2019using,jaiswal2020survey,misra2020self}. The standard SSL methods based on InfoNCE loss \cite{oord2018representation}, such as SimCLR\cite{chen2020simple}, MOCO\cite{he2020momentum} and Self-Con\cite{bae2023self}, has already demonstrated outstanding performance. Further, SupCon\cite{khosla2020supervised} extends contrastive learning to the fully supervised setting.

\cite{vaze2021open} heuristically discovered that a good closed-set accuracy always benefits for open set recognition. Leveraging this viewpoint to the representation learning, \cite{xu2023contrastive} made an initial venture to apply Supervised Contrastive Learning to OSR task, and developed ConOSR which utilizes data augmentation and soft label technologies to representation learning. However, progress in this paradigm is bottlenecked by the absence of a contrastive loss specially designed for pseudo-unknowns.

\subsection{Universum Learning}
The initial concept of universum was introduced by \cite{weston2006inference} as a collection cannot be assigned to any target classes in task. In recent years, \cite{zhang2017universum} extended this to deep learning. \cite{nguyen2017distance} introduced a distance metric learning approach that leverages universum data. \cite{xiao2021new} tackled the challenge of Transductive learning with universum. 

The most recent advancements in this field explored the acquisition of universum through the application of Mixup \cite{zhang2017mixup}. \cite{han2023universum} produced hard negatives in contrastive learning by Mixup-induced universum. \cite{zhang2022class} used \textit{High-order Mixup} for universum to re-balance the classes in long-tailed recognition. In spite of its efficiency, a lack of designing a novel universum suitable for OSR still persists.

\section{Dual Contrastive Learning with Target-Aware Universum}

\subsection{Preliminaries and Problem Statement}

We denote $\mathcal{D_{\mathit{tr}}}$ as a training set consisting  $\mathit{n}$ labeled instances $\left \{  \left ( \boldsymbol{x_{i}},\boldsymbol{y_{i}}  \right )\right \}_{i=1}^{n} $, where $\boldsymbol{y_i}\in $ $\left \{ 1,...,\mathit{K} \right \} $ is the corresponding class label, and $N_{i}$ denotes the number of instances in each class. Additionally, $\mathcal{D_{\mathit{tau}}}$ represents the set of TAU data derived from  $\mathcal{D_{\mathit{tr}}}$, which contains an equal number of $\mathit{n} $ instances$\left \{ \boldsymbol{x_{u}^{i}},\boldsymbol{y_{u}^{i}} \right \}_{i=1}^{n}$, where the labels of pseudo-unknowns $\boldsymbol{y_{u}^{i}}\in $ $\left \{ \mathit{K}+1,...,\mathit{K}+\mathit{K} \right \}$.

We employ a two-step training strategy. In the contrastive learning step, an encoder network $E\left ( \cdot  \right )$ and a projection network $\psi (\cdot )$ are optimized by the contrastive loss based on the features of $\psi (E\left ( \boldsymbol{x_{i}}  \right )) $ and $\psi (E\left ( \boldsymbol{x_{u}^{i}}  \right ))$. In the classifier training step, only $\mathcal{D_{\mathit{tr}}}$ is involved in training, and a classifier $f\left ( \cdot  \right ) $ is optimized by the cross entropy loss. $\mathcal{D_{\mathit{te}}}=\left \{ \boldsymbol{t_{1}},..., \boldsymbol{t_{u}} \right \}_{i=1}^{n} $ denotes a test set which contains the instances drawn from unknown classes.

\subsection{Target-Aware Universum}

Different from existing methods for OSR, we explore the potential to generate universum as the pseudo-unknown data through instances of $K$ known classes straight off the shelf.

\noindent \textbf{Mixup. }Traditional Mixup generates data by linear interpolating pairs of training instances. Given $\boldsymbol{ x_{i}} $ and $\boldsymbol{x_{j}}$, a pair of training instances randomly sampled from  $\mathcal{D_{\mathit{tr}}}$, the synthesized instance is defined as \cite{zhang2017mixup}:
\begin{eqnarray} 
\boldsymbol{\tilde{x}} =\lambda \cdot \boldsymbol{x_{i}}+  \left ( 1-\lambda  \right ) \boldsymbol{x_{j}},
\end{eqnarray} 

\noindent where $\lambda \in \left [ 0,1 \right ]$ is sampled from the Beta distribution. However, it only combines the instance-pairs information of two images which simply captures the \textit{local information} result in generating ambiguous samples. Thus, we introduce a variant of Mixup, namely \textit{Targeted Mixup}, to acquire \textit{global information} among all classes. The outcome is referred to as \textit{Target-Aware Universum} (TAU), which eliminates ambiguous samples to a large extent. As shown in Fig. 3. 

\subsubsection{Definition 1. \textit{Target-Aware Universum.}} 
\textit{Given a subset $\mathcal{D} _{sub}=\left \{ \left (\boldsymbol{x_k},\boldsymbol{y_k} \right) \right \}_{k=1}^{K_B}$ where each instance is randomly sampled from each of the $K_{B}$ known classes in a batch. For $ \forall \boldsymbol{x_{i}} \in \mathcal{D} _{sub}$, its Target-Aware Universum is defined as:}
\begin{eqnarray} 
\boldsymbol{x_{u}^{i}} = \lambda\cdot \boldsymbol{x_{i}} +  \left ( 1- \lambda  \right )\cdot \frac{1}{K_B-1}{\textstyle \sum_{y_j\ne y_i}^{K_{B}-1}} \boldsymbol{x_j},
\end{eqnarray} 

\noindent where ${K_{B}}$ is the number of known classes in a batch.

Compared with the traditional Mixup, besides its ability to eliminate ambiguous samples, the greatest characteristic of Targeted Mixup is targeted known $\boldsymbol{x_i}$ contributes significantly more information to TAU, whereas that of instances in the remaining $K$-$1$ known classes is averaged out. Therefore, TAU could be regarded as the high-quality hard negative \cite{kalantidis2020hard}, as there is a great overlap of semantic information between it and its targeted known. We will discuss why TAU is suitable for $K$+$K$ in OSR in the latter experimental results. 

Further, a more insightful understanding of the essence of \textit{Target-Aware} is that guided by targeted known classes, TAU is endowed with the concept of \textit{independent classes} possessing distinct $K$ classes and $N_{i}$ samples in per class. Consequently, TAU automatically expands the unknown classes from the previous $1$ to $K$ and alleviates the distribution disruption and the imbalance issues on both class-wise and instance-wise.

\subsection{Dual Contrastive Learning Loss}
The contrastive learning loss was introduced to pull an anchor and its positives closer while the negatives are pushed apart\cite{duan2018deep}. Particularly, SupCon loss introduces label information into learning. In this framework, given a training instance $\boldsymbol{x_{i}}$, the network maps it to a representation vector, $\psi (E\left ( \boldsymbol{x_{i}}  \right ))= \boldsymbol{z}_i \in \mathbb{R^{\mathit{D_P}}} $, then the contrastive loss is defined as\cite{khosla2020supervised}:
\begin{eqnarray} 
\mathcal{L}^{sup}=\sum_{i}\frac{-1}{|P(i)|}\sum_{p\in P(i)}\log\frac{\exp{(\boldsymbol{z}_i\cdot \boldsymbol{z}_p/\tau)}}{\sum_{k\ne i}\exp{(\boldsymbol{z}_i\cdot \boldsymbol{z}_k/\tau)}},
\end{eqnarray}
\noindent where $P(i)$ is the set of all positive data in a batch from the class $i$, and $|P(i)|$ is its cardinality. $\tau$ is the scalar temperature hyper-parameter.

\begin{figure}[t]
\centering
\includegraphics[width= 0.9\columnwidth]{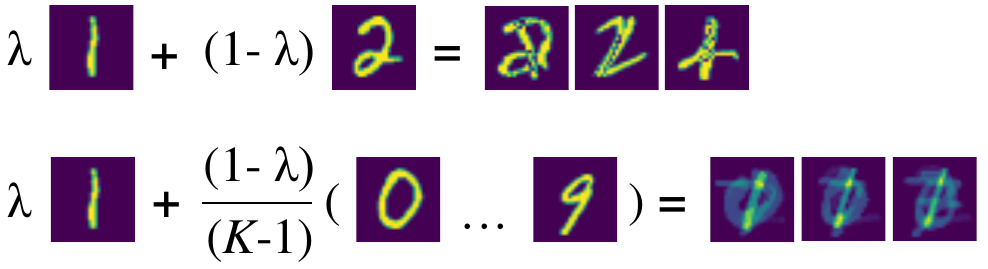} 
\caption{The generated images from mixing digital "1" and "2" may belong to digital "2", "4" and "8" images of known classes(\textit{top}). TAU can highlight the targeted known digital "1" and avoid the \textit{ambiguous samples} (\textit{below}). }
\label{mix}
\end{figure}

With the inclusion of label information, SupCon has resulted in a remarkable improvement in the performance of discrimination. However, given the absence of a dedicated contrastive learning loss designed for unlabeled data, it hinders the comprehensive of unknowns' representation, which unfairly contrasts to unknowns. Thus, we propose an optimized version, namely \textit{Dual Contrastive} (DC) loss, which inherently adapts TAU classes to the same contrastive form of known classes.

The DC loss is composed of two components. In the first part, known classes act as anchors:
\begin{align} 
\mathcal{L}^{k}=\sum_{i\in \mathcal{D}_{tr}}\frac{-1}{|P(i)|}\sum_{p\in P(i)}\log\frac{\exp{(\boldsymbol{z}_i\cdot \boldsymbol{z}_p/\tau)}}{\sum_{k\ne i}\exp{(\boldsymbol{z}_i\cdot \boldsymbol{z}_k/\tau)}+U},
\end{align} 

\noindent where $U = \sum_{\boldsymbol{x^{i}_{u}}\in U(i)}\exp {(\boldsymbol{z_{i}\cdot \boldsymbol{u_{i}}}/\tau)}$,  $U(i) \in $ $\mathcal{D_{\mathit{tau}}}$ represents the set of TAU relative to $i$-th targeted known class, and $\psi (E\left ( \boldsymbol{x_{u}^{i}}  \right ))= \boldsymbol{u}_i \in \mathbb{R^{\mathit{D_P}}} $.

The objective of $\mathcal{L}^{k}$ is to improve the discrimination between the targeted known class and other known classes as well as its TAU class. Each targeted known class acts as an anchor, minimizing the intra-class distance between positively labeled classes while simultaneously maximizing the inter-class margins from negatively labeled classes and its TAU class.

From the insights provided before, we also hope to treat TAU as fairly as known data, allowing them to enhance discriminability through the contrastive learning framework. Then the second part for TAU is defined as:
\begin{align} 
\mathcal{L}^{u}=\sum_{i\in \mathcal{D}_{tau}}\frac{-1}{|U(i)|}\sum_{p\in U(i)}\log\frac{\exp{(\boldsymbol{u}_i\cdot \boldsymbol{u}_p/\tau)}}{\sum_{k\ne i}\exp{(\boldsymbol{u}_i\cdot \boldsymbol{u}_k/\tau)}+K},
\end{align} 

\noindent where $K = \sum_{\boldsymbol{x_{i}}\in P(i)}\exp {(\boldsymbol{u_{i}\cdot \boldsymbol{z_{i}}}/\tau)}$, $P(i) \in$ $\mathcal{D_{\mathit{tr}}}$ is the set of data of the targeted known class relative to $\boldsymbol{x^{i}_{u}}$ and $U(i)$ is the set of all positive data belongs to $\boldsymbol{x^{i}_{u}}$.

$\mathcal{L}^{u}$, as the dual form of $\mathcal{L}^{k}$, differs in the following aspects: (1) anchors are replaced by TAU classes; (2)positively labeled classes are replaced by TAU classes, which share the same label with the anchor $x^{i}_{u}$, boosting a compressed intra-class space among them; (3) negatively labeled classes are replaced by TAU classes with different labels, increasing the inter-class margins; (4) meanwhile, the inter-class distance between $\boldsymbol{x^{i}_{u}}$ and its targeted known class $\boldsymbol{x_{i}}$ is enlarged.

Finally, the loss in the contrastive learning step is a combination:
\begin{eqnarray} 
\mathcal{L}=\mathcal{L}^{k}+\gamma\mathcal{L}^{u},
\end{eqnarray} 

\noindent where $\gamma$ is a balancing coefficient. $\mathcal{L}$ can be regarded as the comprehensive results of fairly contrasting all instances irrespective of known or TAU with their negatives.

\subsection{Theoretical Analysis}
In this subsection, we conduct theoretical analyses from \textit{Fairly Contrast} and \textit{Hard Negative Mining} perspectives. Since $\mathcal{L}^{k}$ and $\mathcal{L}^{u}$ are dual forms of each other, these analyses are similar for both, for simplicity, we consider the $\mathcal{L}^{k}$ as an example. The details can be found in Appendix A.

The gradient for $\mathcal{L}^{k}$ with respect to $\boldsymbol{z}_i$ can be depicted as:

\begin{align}
  & \frac{\partial \mathcal{L}^{k}}{\partial \boldsymbol{z}_i} = -\frac{1}{\tau}\left ( \boldsymbol{z}_p+ \frac{1}{S} {\textstyle\sum_{k\ne i}} C_{k}\cdot \boldsymbol{z}_k + \frac{1}{S} {\textstyle \sum_{\boldsymbol{x^{i}_{u}}\in U(i)}}  O_{i}\cdot \boldsymbol{u}_i \right ) \nonumber \\
  & \quad \quad = -\frac{1}{\tau}\left (  \boldsymbol{z}_p+G_{NK}+G_{TAU} \right ), 
\end{align}

\noindent where $C_{k} = \exp{(\boldsymbol{z}_i\cdot \boldsymbol{z}_k/\tau)}$, $O_{i} = \exp{(\boldsymbol{z}_i\cdot \boldsymbol{u}_i/\tau)}$ and $S = {\textstyle \sum_{k\ne i} C_{k}+\textstyle \sum_{\boldsymbol{x^{i}_{u}}\in U(i)}} O_{i}$. It can be divided into three parts involved in the gradient update process: the representation of positively labeled data, the gradient of negatively labeled known data and the gradient of TAU data.

If TAU is omitted, $\mathcal{L}^{k}$ reverts to $\mathcal{L}^{sup}$, and the gradient for $\mathcal{L}^{sup}$ with respect to $\boldsymbol{z}_i$ will be written as:
\begin{eqnarray} 
\begin{aligned}
\frac{\partial \mathcal{L}^{sup}}{\partial \boldsymbol{z}_i} =-\frac{1}{\tau}  \left (\boldsymbol{z}_p+\frac{ {\textstyle \sum_{k\ne i}} C_{k}\cdot \boldsymbol{z}_k}{ {\textstyle \sum_{k\ne i} C_{k}} }\right ),
\end{aligned}
\end{eqnarray} 

\begin{table*}[htbp]
\centering
\begin{tabular}{c|c|c|c|c|c|c}
\toprule
    Dataset & MNIST & SVHN  & CIFAR10 & CIFAR+10  & CIFAR+50 & {TinyImageNet} \\
\midrule
    Softmax & 97.8  & 88.6  & 67.7  & 81.6  & 80.5  & 57.7 \\
    OpenMax & 98.1  & 89.4  & 69.5  & 81.7  & 79.6  & 57.6 \\
    G-OpenMax & 98.4  & 89.6  & 67.5  & 82.7  & 81.9  & 58.0 \\
    OSRCI & 98.9  & 91.0  & 69.9  & 83.8  & 82.7  & 58.6 \\
    CPN   & 99.0  & 92.6  & 82.8  & 88.1  & 87.9  & 63.9 \\
    RPL++   & 99.3  & 95.1  & 86.1  & 85.6  & 85.0  & 70.2 \\
    GFROSR & -     & 93.5  & 80.7  & 92.8  & 92.6  & 60.8 \\
    PROSER & -     & 94.3  & 89.1  & 96.0  & 95.3  & 69.3 \\
    OpenHybrid & 99.5  & 94.7  & 95.0  & 96.2  & 95.5  & 79.3 \\
    ARPL  & \textbf{99.7}  & 96.7  & 91.0  & 97.1  & 95.1  & 78.2 \\
    Class-inclusion & -  & 95.6  & 94.8  & 96.1  & 95.7  & 78.5 \\
    PMAL & 99.5  & 96.3  & 94.6  & 96.0  & 94.3  & 81.8 \\
    All-U-Need(MLS) & 99.3  & 97.1  & 93.6  & 97.9  & 96.5  & 83.0 \\
    Vanilla SupCon & \textbf{99.7}  & 98.8  & 93.7  & 97.9  & 97.0  & 79.6 \\
    ConOSR & \textbf{99.7}  & 99.1  & 94.2  & 98.1  & 97.3  & 80.9 \\
    \midrule
    DCTAU (w/o DC) & 95.6      & 94.9      & 92.3  & 95.1  & 93.7      & 71.0 \\ 
    DCTAU & \textbf{99.7}  & \textbf{99.2}  & \textbf{95.6}  & \textbf{98.5}  & \textbf{98.1}  & \textbf{83.6} \\
    \bottomrule
    \end{tabular}
\caption{Open Set recognition results in terms of the AUROC(\%). DCTAU (w/o DC) means $\mathcal{L}$ only contains $\mathcal{L}^{k}$. "-" means the original paper does not report the corresponding result. Results are averaged among five randomized trials.}
\label{table1}
\end{table*}%

\noindent \textbf{Fairly Contrast. }Compared with $\partial \mathcal{L}^{sup} /\partial \boldsymbol{z}_i$, the differences in $\partial \mathcal{L}^{k} /\partial \boldsymbol{z}_i$ primarily reflected in: (1) the denominator of $G_{NK}$ contains ${\textstyle \sum_{\boldsymbol{x^{i}_{u}}\in U(i)}} O_{i}$, causing the weight of negatively labeled known data decreases; (2) $G_{TAU}$ added increases the weight of TAU explicitly. In a nutshell, DC loss can inherently adjust the weights between known data and TAU in the gradient update process to pay more attention to unknowns, achieving fairly contrast in OSR.

\noindent \textbf{Hard Negative Mining. }Within Eq. (7), our DC loss can adaptively adjust the weights of $G_{NK}$ and $G_{TAU}$ in which all negatives exist. This adjustment is based on whether the hard negatives attributed to $G_{NK}$ or $G_{TAU}$, which reflects the inner ability of hard negative mining. We further analyze this ability across three situations, and more details will be presented in Appendix A.

\subsection{Rejecting Unknowns}

In this step, a light-weight classifier $f\left ( \cdot  \right ) $ is trained by minimizing the cross entropy loss. 

Given a training instance $\left ( \boldsymbol{x_{i}},\boldsymbol{y_{i}} \right )$, a conventional Softmax classifier outputs the posterior probability of $\boldsymbol{x_{i}}$ belonging to the $k$-th known class by:
\begin{eqnarray} 
\hat{y_{i}}=P(y_{i}=k|\boldsymbol{x_{i}})=\frac{\exp ({f_{k}(E(\boldsymbol{x_{i}})))} }{ {\textstyle \sum_{c=1}^{K}\exp ({f_{c}(E(\boldsymbol{x_{i}})))}} }, 
\end{eqnarray} 

The $f\left ( \cdot  \right )$ is optimized by minimizing the cross entropy loss, $\mathcal{L}^{ce}=-\sum_iy_i\log \hat{y_{i}}$. Followed the  \cite{xu2023contrastive}, the rejection thresholds percentile $\varepsilon _{i}$ are estimated for detecting unknown data.

During the test, the max  posterior probability as confidence score, $\max_{k\in\{1,\cdots,K\}} P(y=k|\boldsymbol{t_{i}})$, and then a test instance $t_{i}$ can be estimated as one of the known classes or recognized as the unknown data by:
\begin{eqnarray} 
\widehat y=\begin{cases}\arg\max_{k\in\{1,\cdots,K\}}P(y=k|\boldsymbol{t_{i}}),&\text{if }conf\geq\varepsilon,\\\text{ unknown class},&\text{otherwise.}\end{cases}
\end{eqnarray} 

\section{Experiments}

\subsection{Experiments for Open Set Recognition}

\textbf{Datasets.} Following the protocol defined in \cite{neal2018open} and the dataset splits with \cite{chen2021adversarial,xu2023contrastive}, a summary of 6 benchmark datasets is provided:

\begin{itemize}
\item \textbf{MNIST,SVHN,CIFAR10.} MNIST\cite{lake2015human}, SVHN\cite{netzer2011reading} and CIFAR10\cite{krizhevsky2009learning} all consist 10 of classes, of which 6 classes are randomly selected as known classes and the other 4 classes as unknown.
\item \textbf{CIFAR+10,CIFAR+50.} In this group of experiments, 4 classes are selected from CIFAR10 as known classes for training, and 10\textbackslash50 classes sampled from CIFAR100\cite{krizhevsky2009learning} as unknown.
\item \textbf{TinyImageNet.} TinyImageNet is a subset derived from ImageNet\cite{russakovsky2015imagenet} consisting of 200 classes. 20 known classes and the left 180 unknown classes are randomly sampled for evaluation.
\end{itemize}

\noindent \textbf{Implementations.} In the contrastive learning step, the feature encoder backbone is the same with \cite{neal2018open}, and an MLP with two fully connected layers is employed as the projection network. In the classifier training step, the network is also an MLP with a 128-node fully connected layer. Followed with \cite{xu2023contrastive}, the training epochs of these two steps are 600 and 20 respectively. More details will be found in Appendix B.

\begin{table*}[t]
\centering
\begin{tabular}{c|c|c|c|c|c|c}
\toprule
   Dataset & MNIST & SVHN  & CIFAR10 & CIFAR+10  & CIFAR+50 & {TinyImageNet} \\
\midrule
    Softmax & 99.2  & 92.8  & 83.8  & 90.9  & 88.5  & 60.8 \\
    GCPL  & 99.1  & 93.4  & 84.3  & 91.0  & 88.3  & 59.3 \\
    RPL   & 99.4  & 93.6  & 85.2  & 91.8  & 89.6  & 53.2 \\
    ARPL  & 99.4  & 94.0  & 86.6  & 93.5  & 91.6  & 62.3 \\
    ARPL+CS & 99.5  & 94.3  & 87.9  & 94.7  & 92.9  & 65.9 \\
    Class-inclusion & -     & 85.4  & 87.0  & 88.1  & 86.5  & 49.3 \\
    \midrule
    DCTAU & \textbf{99.6}  & \textbf{96.2}  & \textbf{93.9}  & \textbf{97.2}  & \textbf{97.1}  & \textbf{77.6} \\
    \bottomrule
    \end{tabular}
\caption{The open set classification rate OSCR(\%) curve results of open set recognition. "-" means the original paper does not report the corresponding result. Results are averaged among five randomized trials.}
\label{table2}
\end{table*}%

\noindent \textbf{Evaluation Metrics. }Area Under the Receiver Operating Characteristic (AUROC) curve\cite{fawcett2006introduction} is used for detecting unknown data from test set. Open Set Classification Rate (OSCR) \cite{dhamija2018reducing,wang2022openauc} is employed for evaluating correct classifications of known classes. Details will be shown in Appendix B.

\noindent \textbf{Results Comparison.} The baselines compared with DCTAU includes Softmax Thresholding\cite{hendrycks2016baseline}, OpenMax\cite{bendale2016towards}, GOpenMax\cite{ge2017generative}, OSRCI\cite{neal2018open}, CPN\cite{yang2020convolutional}, C2AE\cite{oza2019c2ae}, RPL++\cite{chen2020learning}, GFROSR\cite{perera2020generative}, PROSER\cite{zhou2021learning}, OpenHybrid\cite{zhang2020hybrid}, ARPL\cite{chen2021adversarial}, Class-inclusion\cite{cho2022towards}, PMAL\cite{lu2022pmal}, All-U-Need(MLS)\cite{vaze2021open}, and ConOSR\cite{xu2023contrastive}. 

We provide the AUROC performance of different methods in Table 1. DCTAU shows a significant superiority in performance over almost all the methods. Compared to the method employing the contrastive framework, our DCTAU outperforms the recently proposed ConOSR across all datasets, in particular, on the most challenging dataset TinyImageNet by a margin of 2.7\%.

The results of OSCR are shown in Table 2. DCTAU shows a remarkable performance improvement on all datasets. Surprisingly, compared to the second best, the OSCR of DCTAU increased by 11.7\% on TinyImageNet.

\subsection{Experiments for Intrinsic Mechanism of DCTAU}
In this group of experiments, we follow the protocol in \cite{yoshihashi2019classification,zhou2021learning} to explore the intrinsic mechanism for the efficacy of DCTAU through Out-of-distribution detection(OOD). Here, all training classes of the original dataset are used as In-distribution(ID) data. While instances from another dataset are added to the test set as the OOD data.

\noindent \textbf{Datasets.} 
\begin{itemize}
\item \textbf{ID:MNIST/OOD:Omniglot,Noise,MNIST-Noise.} Omniglot\cite{lake2015human} is a dataset of hand-written alphabet characters. Noise is a set of images synthesized by sampling each pixel value from a uniform distribution on [0, 1]. MNIST-Noise is also a synthesized set by superimposing MNIST’s test images on Noise. The number of OOD data is 10,000, equal to the test data of MNIST.

\item \textbf{ID:CIFAR10/OOD:ImageNet,LSUN.} The OOD data is sampled from ImageNet and LSUN\cite{yu2015lsun}. Since the image size of ImageNet and LSUN is different from CIFAR10, we use two different ways to process them. CIFAR10, ImageNet and LSUN all have 10,000 images of their test set. More details are in Appendix C.
\end{itemize}

\begin{table}[t]
  \centering
    \begin{tabular}{c|c|c|c}
    \toprule
    Dataset & Omniglot & MNIST-Noise & Noise \\
    \midrule
    Softmax & 59.5  & 64.1  & 82.9 \\
    OpenMax & 68.0  & 72.0  & 82.6 \\
    CROSR & 79.3  & 82.7  & 82.6 \\
    PRESER & 86.2  & 87.4  & 88.2 \\
    ConSOR & 95.4  & 98.7  & 98.8 \\
        \midrule
    DCTAU & \textbf{96.5}    &  \textbf{99.3}     & \textbf{99.3} \\
    \bottomrule
    \end{tabular}%
  \caption{Out-of-Distribution Detection on MNIST with various datasets added to the test set as unknowns. We report macro F1-scores.}
  \label{table3}%
\end{table}%

\begin{table}[t]
  \centering

    \begin{tabular}{c|c|c|c|c}
    \toprule
    \multirow{2}{*}{Dataset} & ImageNet & ImageNet & LSUN  & LSUN \\
          & (Crop) & (Resize) & (Crop) & (Resize) \\
    \midrule
    Softmax & 63.9  & 65.3  & 64.2  & 64.7 \\
    OpenMax & 66.0    & 68.4  & 65.7  & 66.8 \\
    OSRCI & 63.6  & 63.5  & 65.0    & 64.8 \\
    CROSR & 72.1  & 73.5  & 72.0    & 74.9 \\
    GFROSR & 75.7  & 79.2  & 75.1  & 80.5 \\
    PRESER & 84.9  & 82.4  & 86.7  & 85.6 \\
    ConOSR & 89.1  & 84.3  & 91.2  & 88.1 \\
    \midrule
    DCTAU & \textbf{94.7}  & \textbf{93.2}  & \textbf{95.1}  & \textbf{94.5} \\
    \bottomrule
    \end{tabular}%
 \caption{Out-of-Distribution Detection on CIFAR10 with various datasets added to the test set as unknowns. We report macro F1-scores.}
  \label{table4}%
\end{table}%

\noindent \textbf{Evaluation Metrics.} The macro-averaged F1-scores over all ID and OOD class is used to measure the performance.

\noindent \textbf{Results Comparison. }The results of the ID:MNIST setting are reported in Table 3. DCTAU significantly outperforms other methods. It achieves 99.3\% for detecting noisy OOD images, while 96.5\% in Omniglot dataset for detecting semantic OOD images. The results of the ID:CIFAR10 setting are reported in Table 4. We can see DCTAU also handles these scenarios with the best performance. Especially, for the ImageNet-Resize and LSUN-Resize datasets, DCTAU  excels more than 8.9\% and 6.4\%. Further elaborations about the intrinsic mechanism for the efficacy of DCTAU from two aspects based on \textit{Familiarity Hypothesis} \cite{dietterich2022familiarity} can be found in Appendix C.

In summary, the contrastive learning framework reinforces the grasp of learning the most distinctive information of features among known classes. And our DCTAU enhances this ability, since TAU inherently introduces the distinctiveness to pseudo-unknowns (\textit{i.e., }regards them as $K$ independent classes) and DC loss specifically enhances the contrastive learning for more distinctive information of features of pseudo-unknowns.

\begin{figure}[t]
\centering
\includegraphics[width= 1.0\columnwidth]{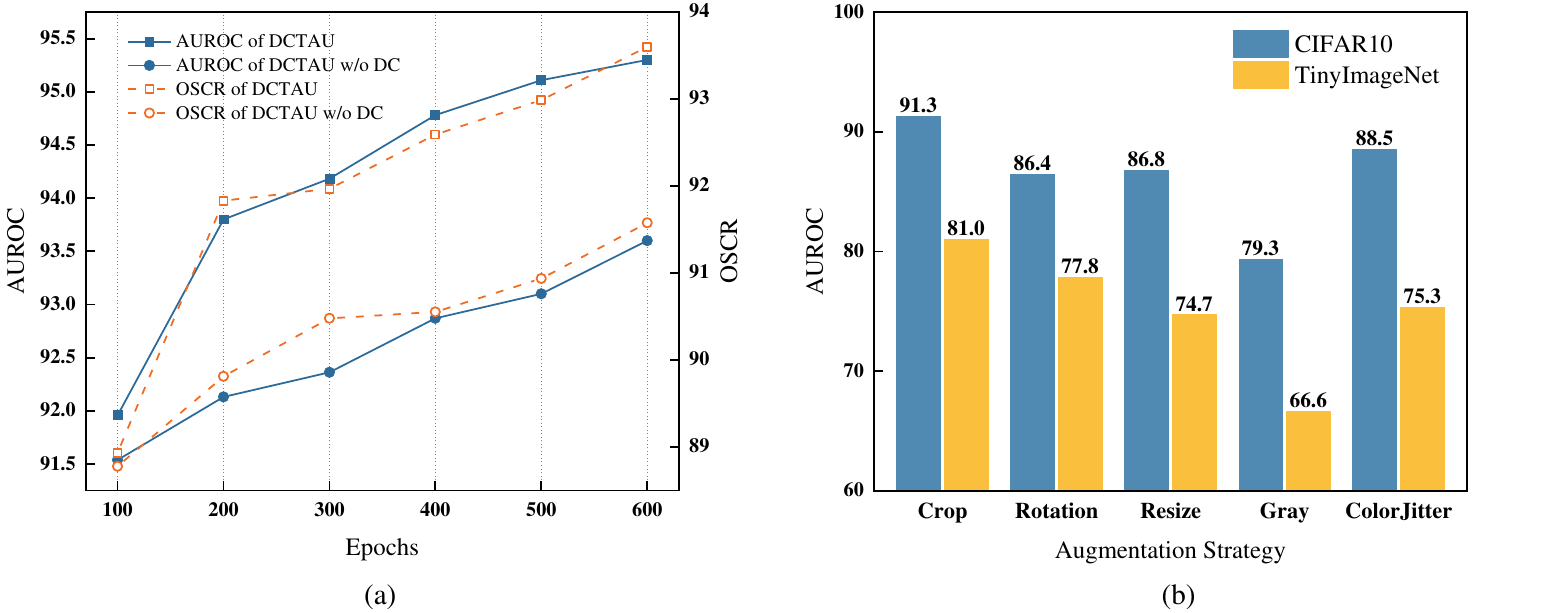} 
\caption{(a) AUROC and OSCR of DCTAU and DCTAU(w/o DC) with varying epochs. The experiments are conducted on CIFAR10; (b) AUROC of DCTAU with varying augmentation techniques. The experiments are conducted on CIFAR10 and TinyImageNet.}
\label{K+1}
\end{figure}

\subsection{Detailed Analysis}
\textbf{Why Target-Aware Universum is Effective. }In this subsection, we perform experiments to shed light on the reasons behind the performance boost attained by employing TAU data as the pseudo-unknowns in the $K$+$K$ strategy.

We conduct a group of experiments on known classes from CIFAR10, to compare the AUROC and OSCR of TAU data with other images including Noise-Gaussian, Noise-Uniform, Natural Images and images generated by traditional Mixup serving as the pseudo-unknown data. The Noise-Gaussian and Noise-Uniform images are the pure noisy images synthesized from a Gaussian distribution and a uniform distribution, respectively. The SVHN dataset is used as the Natural Images and traditional Mixup refers to the images generated based on Eq. (1). The results are shown in Table 5.

\begin{table}[htbp]
  \centering
  \begin{tabular}{c|c|c}
    \toprule
    Method & AUROC & OSCR \\
    \midrule
    Noise-Gaussian & 83.2 & 77.1 \\
    Noise-Uniform & 85.1 & 81.4 \\
    Natural Images & 91.1  & 88.0 \\
     Mixup Images & 93.8 & 92.2 \\
    \midrule
     TAU   & \textbf{95.6}  & \textbf{93.9} \\
    \bottomrule
    \end{tabular}
    \caption{Different data as the Pseudo-unknown employed in the $K$+$K$ strategy.}
  \label{table5}%
\end{table}%

\begin{table}[htbp]
  \centering
    \begin{tabular}{c|ccccc}
    \toprule
    $\lambda$     & 0.1   & 0.3   & 0.5   & 0.7   & 0.9 \\
    \midrule
    AUROC & 93.5  & 94.2  & \textbf{95.6}  & 95.1  & 95.0 \\
    \bottomrule
    \end{tabular}%
      \caption{AUROC under different $\lambda$ for TAU.}
  \label{tab:addlabel}%
\end{table}%

We interpret the results from the perspective of \textit{Hard negative} in contrastive learning. The hard negatives are crucial in learning highly transferable visual representations\cite{kalantidis2020hard}.
The main reason of TAU can be regarded as high-quality hard negatives is that while maintaining better quality of semantic information about their targeted known class, TAU data also possess a certain degree of visual ambiguity. In contrast, Noise-Gaussian and Noise-Uniform guarantee visual ambiguity but are severely limited in the quality of semantic information. Despite the Natural Images can ensure quality, its substantial semantic shift makes them unsuitable as high-quality hard negatives. The decrease in performance of Plain Mixup images is primarily attributed to it may generate ambiguous samples.

\noindent \textbf{Effect of Dual Contrastive loss.} Table 1 presents the results of the performance of DCTAU without $\mathcal{L}^{u}$ which solely emphasizes the discriminative nature of known classes. There is a significant decline of DCTAU (w/o DC) compared to the complete DCTAU, and it even lags behind other baselines. The main reason is the TAU nears to the targeted known class in the feature space. Without a specifically designed contrastive loss to constrain them, they may be easily confused with targeted known classes. This validates the importance of DC loss for our proposed DCTAU framework, and highlights our main viewpoint, that is to handle both known and unknown classes equally in the feature space.

\noindent \textbf{Ablation of Hyper-parameters.} \textbf{(1)About Contrastive Learning.} We modify epochs and the augmentation techniques, both of which are sensitive to contrastive learning. Fig. 4(a) showcases the influence of epochs on CIFAR10. The augmentation technique used in ConOSR is RandAugment\cite{cubuk2020randaugment}. To investigate the impact of different augmentation techniques, we conduct a series of experiments on CIFAR10 and TinyImageNet datasets. The results are shown in Fig. 4(b).  \textbf{(2) $\lambda$ for TAU. }For Targeted Mixup weight $\lambda$, we vary it from 0.1 to 0.9 and 0.5 achieves the best performance. \textbf{(3)Threshold $\epsilon$ and Visualization. } Details about both can be found in Appendix D.

\section{Conclusion}
This paper analyzes the drawbacks of the $K$+$1$ stereotype, and first introduces a $K$+$K$ strategy that emphasizes the comprehension of unknowns during training. Based on this guidance, we propose a novel framework involving two components to achieve all beings are equal in OSR. And extensive experiments on various benchmarks show ours outperforms the state-of-the-art approaches. In future work, we will further explore approaches following the $K$+$K$ strategy and establish the relevant theoretical foundation.

\section{Acknowledgments}
This work was Supported by the National Natural Science Foundation of China (Grant No. 62076124, 62106102), the Natural Science Foundation of Jiangsu Province (Grant No. BK20210292) and the Scientific and technological achievements transferring project of Jiangsu Province (Grant No. BA2021005).

\newpage

\appendix

\section*{\centering\Large Appendix}

\section{A \quad Theoretical Analysis}

\subsection{Gradient Derivation of $\mathcal{L}^{k}$}

\begin{eqnarray} 
\begin{aligned}
\mathcal{L}^{k} = -\frac{\boldsymbol{z}_p}{\tau} \cdot \boldsymbol{z}_i-\log\left [  {\textstyle \sum_{k\ne i}} \exp \left (\boldsymbol{z}_i\cdot  \boldsymbol{z}_k/\tau\right ) \right. \\ \left. + {\textstyle \sum_{\boldsymbol{x^{i}_{u}}\in U(i)}\exp(\boldsymbol{z}_i\cdot  \boldsymbol{u}_i/\tau) } \right ] ,
\end{aligned}
\end{eqnarray} 

where we denote  $C_{k} = \exp{(\boldsymbol{z}_i\cdot \boldsymbol{z}_k/\tau)}$ and $O_{i} = \exp{(\boldsymbol{z}_i\cdot \boldsymbol{u}_i/\tau)}$ and both terms are scalars.

The gradient for $\mathcal{L}^{sup}$ with respect to $\boldsymbol{z}_i$ can be depicted as:
\begin{eqnarray} 
\begin{aligned}
\frac{\partial \mathcal{L}^{sup}}{\partial \boldsymbol{z}_i} =-\frac{1}{\tau} \left (\boldsymbol{z}_p+\frac{ {\textstyle \sum_{k\ne i}} C_{k}\cdot \boldsymbol{z}_k}{ {\textstyle \sum_{k\ne i} C_{k}} }\right ),
\end{aligned}
\end{eqnarray} 

Then the gradient for $\mathcal{L}^{k}$ with respect to $\boldsymbol{z}_i$ can be depicted as:
\begin{align}
  & \frac{\partial \mathcal{L}^{k}}{\partial \boldsymbol{z}_i} = -\frac{1}{\tau}\left ( \boldsymbol{z}_p+ \frac{1}{S} {\textstyle\sum_{k\ne i}} C_{k}\cdot \boldsymbol{z}_k + \frac{1}{S} {\textstyle \sum_{\boldsymbol{x^{i}_{u}}\in U(i)}}  O_{i}\cdot \boldsymbol{u}_i \right ) \nonumber \\
  & \quad \quad = -\frac{1}{\tau}\left (  \boldsymbol{z}_p+G_{NK}+G_{TAU} \right ), 
\end{align}

where $S = {\textstyle \sum_{k\ne i} C_{k}+\textstyle \sum_{\boldsymbol{x^{i}_{u}}\in U(i)}} O_{i}$. The components in $(\cdot)$ can be divided into three parts involved in the gradient update process: the representation of positively labelled data, the gradient of negatively labelled known data and the gradient of TAU data. 

\subsection{Hard Negative Mining}
Analysis regarding \textit{Fairly Contrast} has already been presented in the paper. Here, we will discuss \textit{Hard Negative Mining} across three situations.

\textbf{Situation 1. } When negatively labelled known data is hard and TAU is easy, we have $\boldsymbol{z_{i} \cdot z_{k}} \approx 1$ and $\boldsymbol{z_{i} \cdot u_{i}} \approx 0$, $C_{k} = e^{1/\tau}$, $O_{i} = 1$. Therefore,

\begin{eqnarray} 
\begin{aligned}
\frac{\partial \mathcal{L}^{sup}}{\partial \boldsymbol{z}_i} =-\frac{1}{\tau} \left (\boldsymbol{z}_p+\frac{ {\textstyle \sum_{k\ne i}} e^{1/\tau}\cdot \boldsymbol{z}_k}{ {\textstyle \sum_{k\ne i} e^{1/\tau}} }\right ),
\end{aligned}
\end{eqnarray} 

\begin{align}
\frac{\partial \mathcal{L}^{k}}{\partial \boldsymbol{z}_i} = -\frac{1}{\tau} \left (\boldsymbol{z}_p+ \frac{1}{S_{1}} {\textstyle \sum_{k\ne i}} e^{1/\tau}\cdot \boldsymbol{z}_k + \frac{1}{S_1} {\textstyle \sum_{\boldsymbol{x^{i}_{u}}\in U(i)}}\boldsymbol{u}_i \right ),
\end{align}

where $S_{1} \approx \textstyle \sum_{k\ne i} e^{1/\tau}$. In Eq. (5), the weights of $\boldsymbol{z}_i$ is $e^{1/\tau}$ larger than $\boldsymbol{u}_i$', which enhances the contribution of hard negatives $\boldsymbol{z}_i$ to the gradient update process. Besides, compared with Eq. (4), the additional term $\textstyle \sum_{\boldsymbol{x^{i}_{u}}\in U(i)}\boldsymbol{u}_i$ implies that $\mathcal{L}^{k}$ can take into account the contribution of pseudo-unknowns to gradient updates, which greatly benefits the encoder.

\textbf{Situation 2. } When negatively labelled known data is easy and TAU is hard, we have $\boldsymbol{z_{i} \cdot z_{k}} \approx 0$ and $\boldsymbol{z_{i} \cdot u_{i}} \approx 1$, $C_{k} = 1$, $O_{i} = e^{1/\tau}$. Therefore,

\begin{align}
\frac{\partial \mathcal{L}^{k}}{\partial \boldsymbol{z}_i} = -\frac{1}{\tau} \left (\boldsymbol{z}_p+ \frac{1}{S_{2}} {\textstyle \sum_{k\ne i}} \boldsymbol{z}_k + \frac{1}{S_{2}}{\textstyle \sum_{\boldsymbol{x^{i}_{u}}\in U(i)}} e^{1/\tau}\cdot \boldsymbol{u}_i\right ),
\end{align}

where $S_{2} \approx \textstyle \sum_{\boldsymbol{x^{i}_{u}}\in U(i)} e^{1/\tau}$. It especially enhances the contribution of hard negatives $\boldsymbol{u}_i$ to the gradient updates, which means the inner ability of hard negative mining of ours.

\textbf{Situation 3. } When negatively labelled known data and TAU both are hard, we have $\boldsymbol{z_{i} \cdot z_{k}} \approx 1$ and $\boldsymbol{z_{i} \cdot u_{i}} \approx 1$, $C_{k} = O_{i} = e^{1/\tau}$. Therefore,

\begin{align}
\frac{\partial \mathcal{L}^{k}}{\partial \boldsymbol{z}_i} = -\frac{1}{\tau} \left (\boldsymbol{z}_p+ \frac{1}{S_{3}} {\textstyle \sum_{k\ne i}} e^{1/\tau} \cdot \boldsymbol{z}_k + \frac{1}{S_{3}}{\textstyle \sum_{\boldsymbol{x^{i}_{u}}\in U(i)}} e^{1/\tau}\cdot \boldsymbol{u}_i\right ),
\end{align}

where $S_{3} = \textstyle \sum_{k\ne i} e^{1/\tau} + \textstyle \sum_{\boldsymbol{x^{i}_{u}}\in U(i)} e^{1/\tau}$. In this situation, the contribution of the hard negatives irrespective of negatively labelled known data or TAU data is enhanced.

\section{B \quad More About Experiments }

\subsection{Implementation Details}

All experiments are conducted on a Nvidia RTX 3090 GPU. The parameters of networks are optimized by the Adam optimizer with weight decay = 0.0001 on all datasets except for the TinyImageNet optimized by the SGDM optimizer with momentum = 0.9. The learning rates, strategy of learning rate decay and warm for contrastive learning and classifier training are all followed with \cite{xu2023contrastive}. The batch size of loaded training data is 128 by default. In the data augmentation part, the hyper-parameters of RandAugment are fixed as $N = 1, M = 5$. 

\subsection{Metrics Details}

Area Under the Receiver Operating Characteristic (AUROC) curve, which is widely used in OSR is a threshold-independent metric that measures the probability of a positive example being assigned a higher detection score than a negative example\cite{fawcett2006introduction}. 

Additionally, the goal of OSR task is not limited to only detecting unknown data from test instances that AUROC can evaluate, it also involves considering the accuracy of known classes. Thus, a suitable metric for evaluating correct classifications of known classes, Open Set Classification Rate (OSCR) \cite{dhamija2018reducing}, has been widely used in many recent OSR researches. OSCR  sets the correct classification rate (CCR) in relation to the false positive rate (FPR), where $\delta$ is a probability threshold. Therefore,

\begin{eqnarray}
\begin{aligned}
  & CCR(\delta)\\
  & = \frac{|\{x\in\mathcal{D}_{tr}\wedge\arg max_{k}P(k|\boldsymbol{x})=\hat{k}\wedge P(\hat{k}|x)\geq\delta\}|}{|\mathcal{D}_{tr|}}, 
\end{aligned}
\end{eqnarray}

\begin{align}
FPR(\delta)=\frac{|\{x\in\mathcal{D}_{U}\wedge\arg max_{k}P(k|\boldsymbol{x})\wedge P(\hat{k}|x)\geq\delta\}|}{|\mathcal{D}_{U|}}.
\end{align}

\section{C \quad Experiments for Out-of-Distribution Detection}
\subsection{Details about Datasets}
For X-Crop datasets, the original images are cropped into 32*32. And for X-Resize datasets, the original images are resized into 32*32.

\subsection{Further Elaboration}
The results of these two experiments are completely in accordance with the \textit{Familiarity Hypothesis}\cite{dietterich2022familiarity} that states instead of detecting the novel features presented in an image, the existing model succeeds in OSR task primarily by detecting the absence of familiar features in the image for recognizing the unknowns. 

The difficulty of OOD detection depends on the degree of unfamiliarity (\textit{i.e., }distinction in semantic information within the features) between OOD and ID data. The greater the distinction between ID and OOD data, the more unfamiliar the features of OOD become, leading to easier detection of OOD data.

According to this, we support it through these two experiments from different aspects to modify the \textit{familiarity} between ID and OOD data:

\textbf{Semantic Quality.} In ID:MNIST setting, we vary the semantic quality of OOD data. The performance gap between noisy images (MNIST and MNIST-Noise) and semantic OOD images (Omniglot) can be interpreted as the following. The features of OOD data from the Omniglot dataset possess clear semantic information, which contains much familiar semantic information with MNIST's. Therefore, they can be regarded as the high semantic quality OOD data, which makes detection become more challenging on this. However, OOD data from the MNIST-Noise and Noise datasets exhibit lower semantic quality than Omniglot since the semantic information of the features has been blurred, leading to more unfamiliar features with MNIST.

\textbf{Image Structure.} In ID:CIFAR10 setting, we vary the image structure of OOD data. The performance gap between X-Resize and X-Crop OOD data can be interpreted as the following. X-Resize are full original images resized into 32*32 pixels, which maintains the global semantic structure of images. Thus, X-Resize OOD data contains more familiar features with CIFAR10's, which leads to the detection becoming more challenging on these datasets. Meanwhile, the process of Crop disrupts the global semantic structure of images, which makes more unfamiliar features with CIFAR10. So the performance of X-Crop is more than X-Resize in terms of F1 Scores.

\section{D \quad Ablation of Hyper-parameters}

\subsection{Threshold $\epsilon$}
In this subsection, we explore the impact of different threshold percent $\epsilon$ on macro F1-Scores, which is sensitive to the value of the hyper-parameter $\epsilon$, by adjusting it employed in the experiments for Out-of-Distribution Detection. We vary the value of $\epsilon$ from 1 to 15, and show the macro F1-Scores of two settings and the accuracy of ID and OOD data, respectively.

The results of these groups of experiments are shown in Fig. 1. From Fig. 1(a) and (b) (ID:CIFAR10 Setting), we can observe that with $\epsilon$ increased, the macro F1-Scores of four OOD datasets are increasing simultaneously, even the hardest dataset, ImageNet-Resize, can also achieve around 95\% ultimately. The trend begins to stabilize when $\epsilon$ is set to 5. The accuracy of four OOD datasets. The pattern of the accuracy of OOD data changing with $\epsilon$ matches that of the macro F1-Scores. Meanwhile, in our framework, the impact of accuracy on ID data is not significant, consistently staying around 94.8\%. 

In Fig. 1(c) and (d) (ID:MNIST Setting), the hardest OOD dataset is Omniglot and its macro F1-Scores increases with $\epsilon$ increased and is steady when $\epsilon$ is 5. The macro F1-Scores of MNIST-Noise is also stable when $\epsilon$ is 5. Especially, the macro F1-Scores of Noise data reaches almost 100\% right from the beginning and consistently maintained it. The accuracy of these keeps the same trend as that of macro F1-Scores.

From Fig. 1, considering both macro F1-Scores and accuracy together, we set the value of threshold percentile $\epsilon$ as 5 in these experiments.

\subsection{Visualization}
We compare the results of experiments including $K$+$1$ stereotype, DCTAU (w/o DC) and complete DCATU ($K$+$K$) for Open Set Recognition on MNIST dataset. The results shown in Fig. 2 validates the effectiveness of $K$+$K$ pattern and our DCTAU from the visualization.

\begin{figure*}
\centering
\includegraphics[width=2.05\columnwidth]{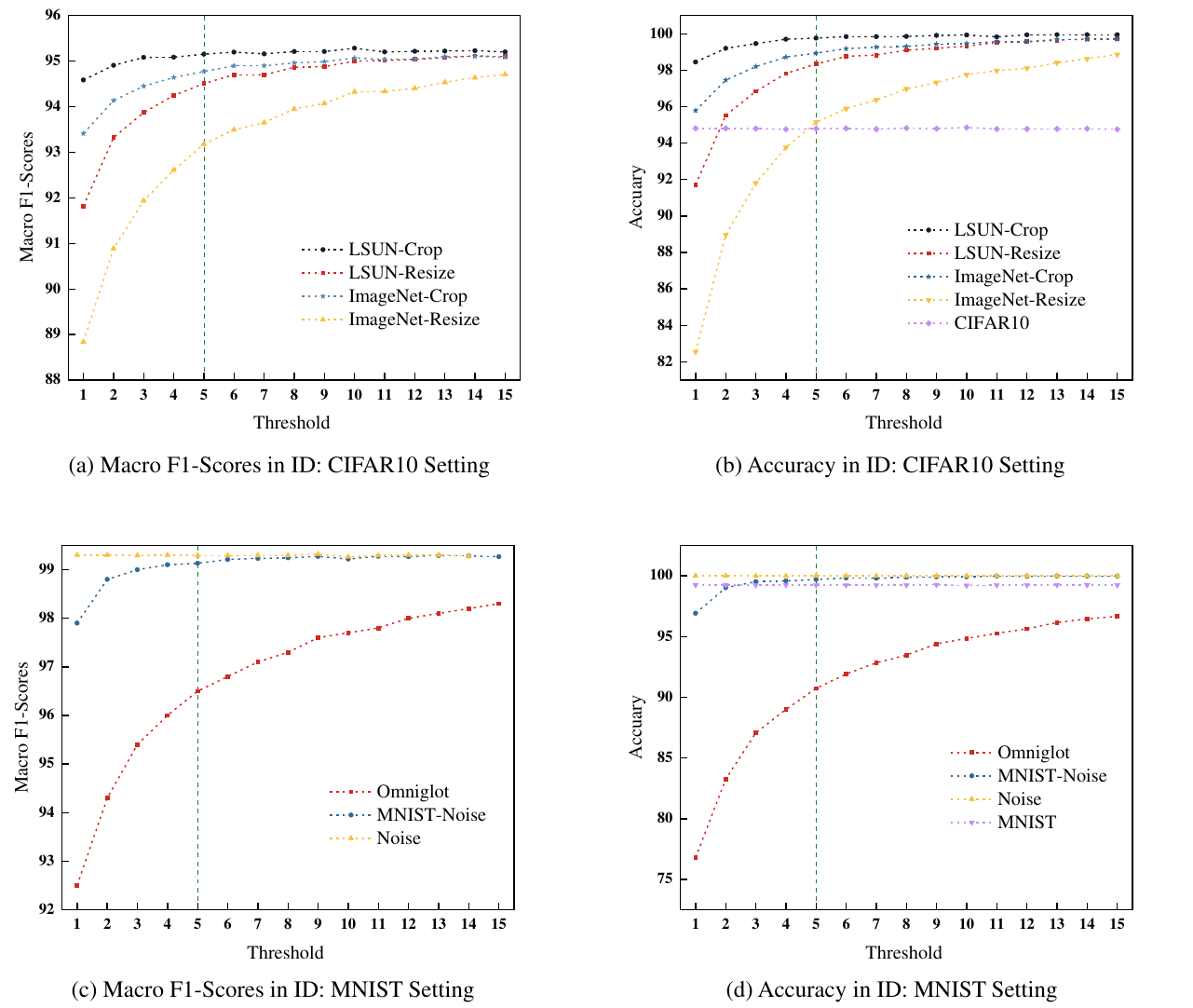} 
\caption{Ablation of Hyper-parameters $\epsilon$.}
\label{fig1}
\end{figure*}

\begin{figure*}
\centering
\includegraphics[width=2.05\columnwidth]{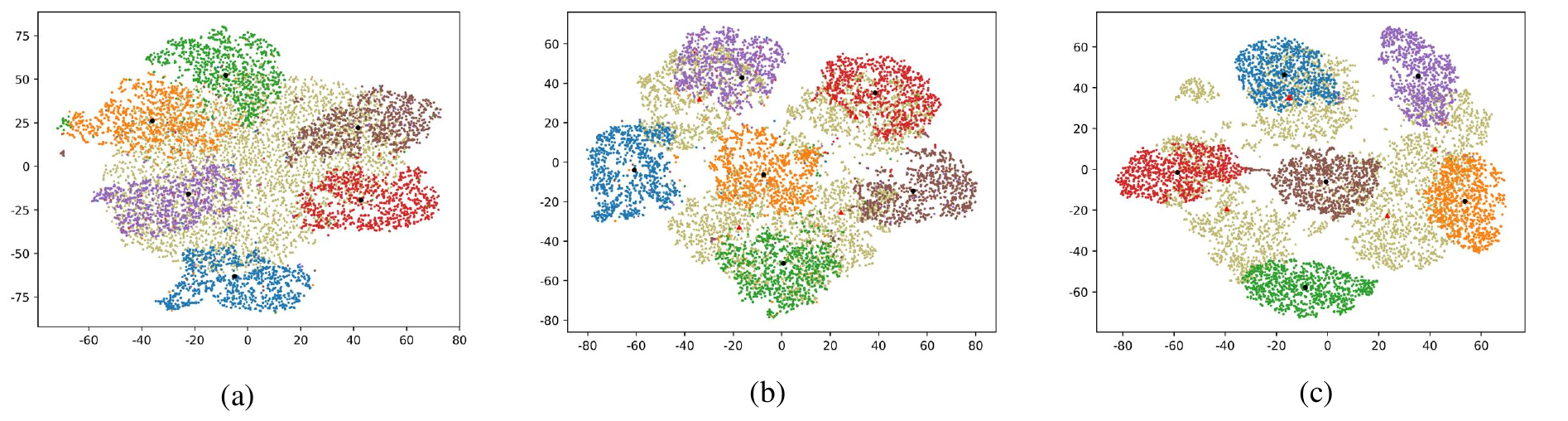} 
\caption{Visualizations of (a) $K$+$1$ stereotype, (b) DCTAU (w/o DC) and DCATU ($K$+$K$). Unknown clasees are the blackish green point. The black circles are the centers of known classes and the red triangles are the centers of unknown classes.}
\label{fig2}
\end{figure*}

\bigskip


\begin{thebibliography}{52}
\providecommand{\natexlab}[1]{#1}

\bibitem[{Bae et~al.(2023)Bae, Kim, Ko, Lee, Noh, and Yun}]{bae2023self}
Bae, S.; Kim, S.; Ko, J.; Lee, G.; Noh, S.; and Yun, S.-Y. 2023.
\newblock Self-Contrastive Learning: Single-Viewed Supervised Contrastive Framework Using Sub-network.
\newblock In \emph{Proceedings of the AAAI Conference on Artificial Intelligence}, volume~37, 197--205.

\bibitem[{Bendale and Boult(2016)}]{bendale2016towards}
Bendale, A.; and Boult, T.~E. 2016.
\newblock Towards open set deep networks.
\newblock In \emph{Proceedings of the IEEE conference on computer vision and pattern recognition}, 1563--1572.

\bibitem[{Chapelle et~al.(2007)Chapelle, Agarwal, Sinz, and Sch{\"o}lkopf}]{chapelle2007analysis}
Chapelle, O.; Agarwal, A.; Sinz, F.; and Sch{\"o}lkopf, B. 2007.
\newblock An analysis of inference with the universum.
\newblock \emph{Advances in neural information processing systems}, 20.

\bibitem[{Chen et~al.(2021)Chen, Peng, Wang, and Tian}]{chen2021adversarial}
Chen, G.; Peng, P.; Wang, X.; and Tian, Y. 2021.
\newblock Adversarial reciprocal points learning for open set recognition.
\newblock \emph{IEEE Transactions on Pattern Analysis and Machine Intelligence}, 44(11): 8065--8081.

\bibitem[{Chen et~al.(2020{\natexlab{a}})Chen, Qiao, Shi, Peng, Li, Huang, Pu, and Tian}]{chen2020learning}
Chen, G.; Qiao, L.; Shi, Y.; Peng, P.; Li, J.; Huang, T.; Pu, S.; and Tian, Y. 2020{\natexlab{a}}.
\newblock Learning open set network with discriminative reciprocal points.
\newblock In \emph{Computer Vision--ECCV 2020: 16th European Conference, Glasgow, UK, August 23--28, 2020, Proceedings, Part III 16}, 507--522. Springer.

\bibitem[{Chen et~al.(2020{\natexlab{b}})Chen, Kornblith, Norouzi, and Hinton}]{chen2020simple}
Chen, T.; Kornblith, S.; Norouzi, M.; and Hinton, G. 2020{\natexlab{b}}.
\newblock A simple framework for contrastive learning of visual representations.
\newblock In \emph{International conference on machine learning}, 1597--1607. PMLR.

\bibitem[{Cho and Choo(2022)}]{cho2022towards}
Cho, W.; and Choo, J. 2022.
\newblock Towards accurate open-set recognition via background-class regularization.
\newblock In \emph{European Conference on Computer Vision}, 658--674. Springer.

\bibitem[{Cubuk et~al.(2020)Cubuk, Zoph, Shlens, and Le}]{cubuk2020randaugment}
Cubuk, E.~D.; Zoph, B.; Shlens, J.; and Le, Q.~V. 2020.
\newblock Randaugment: Practical automated data augmentation with a reduced search space.
\newblock In \emph{Proceedings of the IEEE/CVF conference on computer vision and pattern recognition workshops}, 702--703.

\bibitem[{Dhamija, G{\"u}nther, and Boult(2018)}]{dhamija2018reducing}
Dhamija, A.~R.; G{\"u}nther, M.; and Boult, T. 2018.
\newblock Reducing network agnostophobia.
\newblock \emph{Advances in Neural Information Processing Systems}, 31.

\bibitem[{Dietterich and Guyer(2022)}]{dietterich2022familiarity}
Dietterich, T.~G.; and Guyer, A. 2022.
\newblock The familiarity hypothesis: Explaining the behavior of deep open set methods.
\newblock \emph{Pattern Recognition}, 132: 108931.

\bibitem[{Du et~al.(2022)Du, Wang, Cai, and Li}]{du2022vos}
Du, X.; Wang, Z.; Cai, M.; and Li, Y. 2022.
\newblock Vos: Learning what you don't know by virtual outlier synthesis.
\newblock \emph{arXiv preprint arXiv:2202.01197}.

\bibitem[{Duan et~al.(2018)Duan, Zheng, Lin, Lu, and Zhou}]{duan2018deep}
Duan, Y.; Zheng, W.; Lin, X.; Lu, J.; and Zhou, J. 2018.
\newblock Deep adversarial metric learning.
\newblock In \emph{Proceedings of the IEEE Conference on Computer Vision and Pattern Recognition}, 2780--2789.

\bibitem[{Fawcett(2006)}]{fawcett2006introduction}
Fawcett, T. 2006.
\newblock An introduction to ROC analysis.
\newblock \emph{Pattern recognition letters}, 27(8): 861--874.

\bibitem[{Ge et~al.(2017)Ge, Demyanov, Chen, and Garnavi}]{ge2017generative}
Ge, Z.; Demyanov, S.; Chen, Z.; and Garnavi, R. 2017.
\newblock Generative openmax for multi-class open set classification.
\newblock \emph{arXiv preprint arXiv:1707.07418}.

\bibitem[{Geng, Huang, and Chen(2020)}]{geng2020recent}
Geng, C.; Huang, S.-j.; and Chen, S. 2020.
\newblock Recent advances in open set recognition: A survey.
\newblock \emph{IEEE transactions on pattern analysis and machine intelligence}, 43(10): 3614--3631.

\bibitem[{Gunther et~al.(2017)Gunther, Cruz, Rudd, and Boult}]{gunther2017toward}
Gunther, M.; Cruz, S.; Rudd, E.~M.; and Boult, T.~E. 2017.
\newblock Toward open-set face recognition.
\newblock In \emph{Proceedings of the IEEE Conference on Computer Vision and Pattern Recognition Workshops}, 71--80.

\bibitem[{Han, Geng, and Chen(2023)}]{han2023universum}
Han, A.; Geng, C.; and Chen, S. 2023.
\newblock Universum-inspired Supervised Contrastive Learning.
\newblock \emph{IEEE Transactions on Image Processing}.

\bibitem[{He et~al.(2020)He, Fan, Wu, Xie, and Girshick}]{he2020momentum}
He, K.; Fan, H.; Wu, Y.; Xie, S.; and Girshick, R. 2020.
\newblock Momentum contrast for unsupervised visual representation learning.
\newblock In \emph{Proceedings of the IEEE/CVF conference on computer vision and pattern recognition}, 9729--9738.

\bibitem[{Hendrycks and Gimpel(2016)}]{hendrycks2016baseline}
Hendrycks, D.; and Gimpel, K. 2016.
\newblock A baseline for detecting misclassified and out-of-distribution examples in neural networks.
\newblock \emph{arXiv preprint arXiv:1610.02136}.

\bibitem[{Hendrycks et~al.(2019)Hendrycks, Mazeika, Kadavath, and Song}]{hendrycks2019using}
Hendrycks, D.; Mazeika, M.; Kadavath, S.; and Song, D. 2019.
\newblock Using self-supervised learning can improve model robustness and uncertainty.
\newblock \emph{Advances in neural information processing systems}, 32.

\bibitem[{Jaiswal et~al.(2020)Jaiswal, Babu, Zadeh, Banerjee, and Makedon}]{jaiswal2020survey}
Jaiswal, A.; Babu, A.~R.; Zadeh, M.~Z.; Banerjee, D.; and Makedon, F. 2020.
\newblock A survey on contrastive self-supervised learning.
\newblock \emph{Technologies}, 9(1): 2.

\bibitem[{Kalantidis et~al.(2020)Kalantidis, Sariyildiz, Pion, Weinzaepfel, and Larlus}]{kalantidis2020hard}
Kalantidis, Y.; Sariyildiz, M.~B.; Pion, N.; Weinzaepfel, P.; and Larlus, D. 2020.
\newblock Hard negative mixing for contrastive learning.
\newblock \emph{Advances in Neural Information Processing Systems}, 33: 21798--21809.

\bibitem[{Khosla et~al.(2020)Khosla, Teterwak, Wang, Sarna, Tian, Isola, Maschinot, Liu, and Krishnan}]{khosla2020supervised}
Khosla, P.; Teterwak, P.; Wang, C.; Sarna, A.; Tian, Y.; Isola, P.; Maschinot, A.; Liu, C.; and Krishnan, D. 2020.
\newblock Supervised contrastive learning.
\newblock \emph{Advances in neural information processing systems}, 33: 18661--18673.

\bibitem[{Kong and Ramanan(2021)}]{kong2021opengan}
Kong, S.; and Ramanan, D. 2021.
\newblock Opengan: Open-set recognition via open data generation.
\newblock In \emph{Proceedings of the IEEE/CVF International Conference on Computer Vision}, 813--822.

\bibitem[{Krizhevsky, Hinton et~al.(2009)}]{krizhevsky2009learning}
Krizhevsky, A.; Hinton, G.; et~al. 2009.
\newblock Learning multiple layers of features from tiny images.

\bibitem[{Lake, Salakhutdinov, and Tenenbaum(2015)}]{lake2015human}
Lake, B.~M.; Salakhutdinov, R.; and Tenenbaum, J.~B. 2015.
\newblock Human-level concept learning through probabilistic program induction.
\newblock \emph{Science}, 350(6266): 1332--1338.

\bibitem[{Lu et~al.(2022)Lu, Xu, Li, Cheng, and Niu}]{lu2022pmal}
Lu, J.; Xu, Y.; Li, H.; Cheng, Z.; and Niu, Y. 2022.
\newblock Pmal: Open set recognition via robust prototype mining.
\newblock In \emph{Proceedings of the AAAI Conference on Artificial Intelligence}, volume~36, 1872--1880.

\bibitem[{Misra and Maaten(2020)}]{misra2020self}
Misra, I.; and Maaten, L. v.~d. 2020.
\newblock Self-supervised learning of pretext-invariant representations.
\newblock In \emph{Proceedings of the IEEE/CVF conference on computer vision and pattern recognition}, 6707--6717.

\bibitem[{Neal et~al.(2018)Neal, Olson, Fern, Wong, and Li}]{neal2018open}
Neal, L.; Olson, M.; Fern, X.; Wong, W.-K.; and Li, F. 2018.
\newblock Open set learning with counterfactual images.
\newblock In \emph{Proceedings of the European Conference on Computer Vision (ECCV)}, 613--628.

\bibitem[{Netzer et~al.(2011)Netzer, Wang, Coates, Bissacco, Wu, and Ng}]{netzer2011reading}
Netzer, Y.; Wang, T.; Coates, A.; Bissacco, A.; Wu, B.; and Ng, A.~Y. 2011.
\newblock Reading digits in natural images with unsupervised feature learning.

\bibitem[{Nguyen, Morell, and De~Baets(2017)}]{nguyen2017distance}
Nguyen, B.; Morell, C.; and De~Baets, B. 2017.
\newblock Distance metric learning with the Universum.
\newblock \emph{Pattern Recognition Letters}, 100: 37--43.

\bibitem[{Oord, Li, and Vinyals(2018)}]{oord2018representation}
Oord, A. v.~d.; Li, Y.; and Vinyals, O. 2018.
\newblock Representation learning with contrastive predictive coding.
\newblock \emph{arXiv preprint arXiv:1807.03748}.

\bibitem[{Oza and Patel(2019)}]{oza2019c2ae}
Oza, P.; and Patel, V.~M. 2019.
\newblock C2ae: Class conditioned auto-encoder for open-set recognition.
\newblock In \emph{Proceedings of the IEEE/CVF Conference on Computer Vision and Pattern Recognition}, 2307--2316.

\bibitem[{Perera et~al.(2020)Perera, Morariu, Jain, Manjunatha, Wigington, Ordonez, and Patel}]{perera2020generative}
Perera, P.; Morariu, V.~I.; Jain, R.; Manjunatha, V.; Wigington, C.; Ordonez, V.; and Patel, V.~M. 2020.
\newblock Generative-discriminative feature representations for open-set recognition.
\newblock In \emph{Proceedings of the IEEE/CVF Conference on Computer Vision and Pattern Recognition}, 11814--11823.

\bibitem[{Perera and Patel(2019)}]{perera2019deep}
Perera, P.; and Patel, V.~M. 2019.
\newblock Deep transfer learning for multiple class novelty detection.
\newblock In \emph{Proceedings of the ieee/cvf conference on computer vision and pattern recognition}, 11544--11552.

\bibitem[{Russakovsky et~al.(2015)Russakovsky, Deng, Su, Krause, Satheesh, Ma, Huang, Karpathy, Khosla, Bernstein et~al.}]{russakovsky2015imagenet}
Russakovsky, O.; Deng, J.; Su, H.; Krause, J.; Satheesh, S.; Ma, S.; Huang, Z.; Karpathy, A.; Khosla, A.; Bernstein, M.; et~al. 2015.
\newblock Imagenet large scale visual recognition challenge.
\newblock \emph{International journal of computer vision}, 115: 211--252.

\bibitem[{Scheirer et~al.(2012)Scheirer, de~Rezende~Rocha, Sapkota, and Boult}]{scheirer2012toward}
Scheirer, W.~J.; de~Rezende~Rocha, A.; Sapkota, A.; and Boult, T.~E. 2012.
\newblock Toward open set recognition.
\newblock \emph{IEEE transactions on pattern analysis and machine intelligence}, 35(7): 1757--1772.

\bibitem[{Vaze et~al.(2021)Vaze, Han, Vedaldi, and Zisserman}]{vaze2021open}
Vaze, S.; Han, K.; Vedaldi, A.; and Zisserman, A. 2021.
\newblock Open-set recognition: A good closed-set classifier is all you need?
\newblock \emph{arXiv preprint arXiv:2110.06207}.

\bibitem[{Verma et~al.(2019)Verma, Lamb, Beckham, Najafi, Mitliagkas, Lopez-Paz, and Bengio}]{verma2019manifold}
Verma, V.; Lamb, A.; Beckham, C.; Najafi, A.; Mitliagkas, I.; Lopez-Paz, D.; and Bengio, Y. 2019.
\newblock Manifold mixup: Better representations by interpolating hidden states.
\newblock In \emph{International conference on machine learning}, 6438--6447. PMLR.

\bibitem[{Wang et~al.(2022)Wang, Xu, Yang, He, Cao, and Huang}]{wang2022openauc}
Wang, Z.; Xu, Q.; Yang, Z.; He, Y.; Cao, X.; and Huang, Q. 2022.
\newblock OpenAUC: Towards AUC-Oriented Open-Set Recognition.
\newblock \emph{Advances in Neural Information Processing Systems}, 35: 25033--25045.

\bibitem[{Weston et~al.(2006)Weston, Collobert, Sinz, Bottou, and Vapnik}]{weston2006inference}
Weston, J.; Collobert, R.; Sinz, F.; Bottou, L.; and Vapnik, V. 2006.
\newblock Inference with the universum.
\newblock In \emph{Proceedings of the 23rd international conference on Machine learning}, 1009--1016.

\bibitem[{Xiao, Feng, and Liu(2021)}]{xiao2021new}
Xiao, Y.; Feng, J.; and Liu, B. 2021.
\newblock A new transductive learning method with universum data.
\newblock \emph{Applied Intelligence}, 1--13.

\bibitem[{Xu, Shen, and Zhao(2023)}]{xu2023contrastive}
Xu, B.; Shen, F.; and Zhao, J. 2023.
\newblock Contrastive Open Set Recognition.
\newblock In \emph{Proceedings of the AAAI Conference on Artificial Intelligence}, volume~37, 10546--10556.

\bibitem[{Yang et~al.(2020)Yang, Zhang, Yin, Yang, and Liu}]{yang2020convolutional}
Yang, H.-M.; Zhang, X.-Y.; Yin, F.; Yang, Q.; and Liu, C.-L. 2020.
\newblock Convolutional prototype network for open set recognition.
\newblock \emph{IEEE Transactions on Pattern Analysis and Machine Intelligence}, 44(5): 2358--2370.

\bibitem[{Yoshihashi et~al.(2019)Yoshihashi, Shao, Kawakami, You, Iida, and Naemura}]{yoshihashi2019classification}
Yoshihashi, R.; Shao, W.; Kawakami, R.; You, S.; Iida, M.; and Naemura, T. 2019.
\newblock Classification-reconstruction learning for open-set recognition.
\newblock In \emph{Proceedings of the IEEE/CVF Conference on Computer Vision and Pattern Recognition}, 4016--4025.

\bibitem[{Yu et~al.(2015)Yu, Seff, Zhang, Song, Funkhouser, and Xiao}]{yu2015lsun}
Yu, F.; Seff, A.; Zhang, Y.; Song, S.; Funkhouser, T.; and Xiao, J. 2015.
\newblock Lsun: Construction of a large-scale image dataset using deep learning with humans in the loop.
\newblock \emph{arXiv preprint arXiv:1506.03365}.

\bibitem[{Zhang, Geng, and Chen(2022)}]{zhang2022class}
Zhang, E.; Geng, C.; and Chen, S. 2022.
\newblock Class-Aware Universum Inspired Re-Balance Learning for Long-Tailed Recognition.
\newblock \emph{arXiv preprint arXiv:2207.12808}.

\bibitem[{Zhang et~al.(2017)Zhang, Cisse, Dauphin, and Lopez-Paz}]{zhang2017mixup}
Zhang, H.; Cisse, M.; Dauphin, Y.~N.; and Lopez-Paz, D. 2017.
\newblock mixup: Beyond empirical risk minimization.
\newblock \emph{arXiv preprint arXiv:1710.09412}.

\bibitem[{Zhang et~al.(2020)Zhang, Li, Guo, and Guo}]{zhang2020hybrid}
Zhang, H.; Li, A.; Guo, J.; and Guo, Y. 2020.
\newblock Hybrid models for open set recognition.
\newblock In \emph{Computer Vision--ECCV 2020: 16th European Conference, Glasgow, UK, August 23--28, 2020, Proceedings, Part III 16}, 102--117. Springer.

\bibitem[{Zhang and LeCun(2017)}]{zhang2017universum}
Zhang, X.; and LeCun, Y. 2017.
\newblock Universum prescription: Regularization using unlabeled data.
\newblock In \emph{Proceedings of the AAAI Conference on Artificial Intelligence}, volume~31.

\bibitem[{Zhou, Ye, and Zhan(2021)}]{zhou2021learning}
Zhou, D.-W.; Ye, H.-J.; and Zhan, D.-C. 2021.
\newblock Learning placeholders for open-set recognition.
\newblock In \emph{Proceedings of the IEEE/CVF conference on computer vision and pattern recognition}, 4401--4410.

\bibitem[{Zhu et~al.(2023)Zhu, Cheng, Zhang, and Liu}]{zhu2023openmix}
Zhu, F.; Cheng, Z.; Zhang, X.-Y.; and Liu, C.-L. 2023.
\newblock OpenMix: Exploring Outlier Samples for Misclassification Detection.
\newblock In \emph{Proceedings of the IEEE/CVF Conference on Computer Vision and Pattern Recognition}, 12074--12083.

\end{thebibliography}
\end{document}